\documentclass[conference]{IEEEtran}
\IEEEoverridecommandlockouts
\usepackage{cite}
\usepackage{amsmath,amssymb,amsfonts}
\usepackage{algorithmic}
\usepackage{float} 
\usepackage{graphicx}
\usepackage{hyperref}
\usepackage{textcomp}
\usepackage{xcolor}
\def\BibTeX{{\rm B\kern-.05em{\sc i\kern-.025em b}\kern-.08em
    T\kern-.1667em\lower.7ex\hbox{E}\kern-.125emX}}

\usepackage{amsmath}
\usepackage{amsfonts}
\usepackage{booktabs}
\usepackage{tikz}
\usetikzlibrary{shapes, arrows, positioning, quotes, arrows.meta, decorations.markings}
\usepackage{pgfplots}

\begin{document}

\title{Using Connectome Features to Constrain Echo State Networks}

\author{\IEEEauthorblockN{1\textsuperscript{st} Jacob Morra}
\IEEEauthorblockA{\textit{Department of Computer Science} \\
\textit{The University of Western Ontario}\\
London, ON Canada \\
jmorra6@uwo.ca}
\and

\IEEEauthorblockN{2\textsuperscript{nd} Mark Daley}
\IEEEauthorblockA{\textit{Department of Computer Science} \\
\textit{The University of Western Ontario}\\
London, ON Canada \\
mdaley2@uwo.ca}
}

\maketitle
\begin{abstract}
We report an improvement to the conventional Echo State Network (ESN) across three benchmark chaotic time-series prediction tasks using fruit fly connectome data alone. We also investigate the impact of key connectome-derived structural features on prediction performance – uniquely bridging neurobiological structure and machine learning function; and find that both increasing the global average clustering coefficient and modifying the position of weights – by permuting their synapse-synapse partners – can lead to increased model variance and (in some cases) degraded performance. In all we consider four topological point modifications to a connectome-derived ESN reservoir (null model): namely, we alter the network sparsity, re-draw nonzero weights from a uniform distribution, permute nonzero weight positions, and increase the network global average clustering coefficient. We compare the four resulting ESN model classes – and the null model – with a conventional ESN by conducting time-series prediction experiments on size-variants of the Mackey-Glass 17 (MG-17), Lorenz, and Rossler chaotic time series; denoting each model’s performance and variance across train-validate trials.
\end{abstract}

\begin{IEEEkeywords}
structured reservoir computing, chaotic time series prediction, brain-inspired machine learning, connectome topology, echo state networks
\end{IEEEkeywords}

\vspace{.5cm}\section{Introduction}\vspace{.5cm}

Reservoir Computers (RCs) have grown to occupy a particular niche as a class of machine learning models which are best-in-class in time series prediction of dynamical and chaotic systems \cite{Gauthier2021-yv}, whilst also being efficient and easy to train compared to many other Recurrent Neural Network paradigms. The \textit{trick} to the RC's efficiency is in its \textit{randomly-generated} reservoir layer. Instead of optimizing the weights in this layer, the RC relies on a sufficiently-large population of neurons which are sparsely and randomly connected; this provides the network with a capacity to generate rich transformations of the input into activation states \cite{Gallicchio2020-fl} -- the network is then able to generate a particular output realization by taking a linear combination of these.
\\

What makes the RC shine is also a hindrance, however. \textit{Randomness}, as mentioned, is \textit{non-optimal} by definition. Random weight generation in particular also contributes to large model variance -- for which RCs are notorious \cite{Wu2018-ey}. While, intuitively, one might consider moving to a larger reservoir size to help compensate for ``bad'' random weights, this approach has actually been shown to \textit{decrease} the likelihood that an \textit{effective} network will be found \cite{Koryakin2012-ua}.
\\

Many recent works have sought to counteract the shortcomings of the random RC reservoir. \cite{Gauthier2021-yv}, for example, introduces the Next-Generation Reservoir Computer (NGRC), which replaces the reservoir layer altogether with a time-delayed input sequence, paired with nonlinear functional mappings of the sequence components -- culminating in an output layer which concatenates the linear and nonlinear components; \cite{Gauthier2021-yv} reports similar performance to the conventional RC, but with a tens to hundred-fold decrease in computational complexity. Another recent trend -- coined formally in \cite{Dominey2022} -- is ``Structured Reservoir Computing'', which asserts topological elements or rules onto the reservoir. In \cite{Dominey2022}, for example, authors enforce an exponential distance rule in their reservoir -- inspired by white matter connectivity in \\

\begin{figure}[H]
\centering
{
\begin{tikzpicture}[thick, scale=0.33]
    \begin{axis}[%
        hide axis,
        domain = 0:15,
        samples = 16]
        \addplot+[ycomb,blue,thick] {sin(2*180*x/13)};
    \end{axis}
\end{tikzpicture}

\begin{tikzpicture}[font=\small,thick]
\node[draw,
    align=center,
    rounded rectangle,
    minimum width=2.5cm,
    fill=violet!30,
    minimum height=1cm] (input) {input};

\node[draw,
    above=0.1cm of input,
    opacity=0,
    text width=1.8cm,
    align=center,
    minimum width=1cm,
    fill=cyan!30,
    minimum height=0.7cm,        
    inner sep=0] (input_ts) {\textbf{}};

\node[draw,
    yshift=-3cm,
    above=0.1cm of input,
    opacity=0,
    text width=1.8cm,
    align=center,
    minimum width=1cm,
    fill=cyan!30,
    minimum height=1.5cm,        
    inner sep=0] (input_ts2) {\textbf{}};

\node[draw,
    below=of input,
    minimum width=2.5cm,
    rounded rectangle,
    fill=magenta!30,
    minimum height=1cm,
    align=center,
    inner sep=0] (reservoir) {reservoir};
\node[draw,
    rounded rectangle,
    below=of reservoir,
    minimum width=2.5cm,
    fill=purple!30,
    minimum height=1cm,
    align=center,
    inner sep=0] (output) {output};

\draw (input) edge[->,"$\mathbf{W}^{in}$"] (reservoir);
\draw (reservoir) edge[->,densely dotted,"$\mathbf{W}^{out}$"] (output);
\draw [->] (reservoir) edge[loop left, looseness=8]node{$\mathbf{W}$} (reservoir);
\draw [->] (reservoir) edge[loop right,opacity=0.0]node{} (reservoir);
\draw (input_ts.north west) edge[blue, dashed] (input.north);
\draw (input_ts.north east) edge[blue, dashed] (input.north);
\end{tikzpicture}
}
\caption{The Echo State Network (ESN).}
\label{esn}

\end{figure}
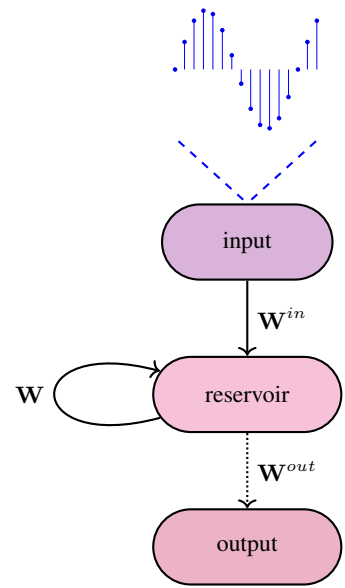

\noindent posterior-anterior cortical hierarchy -- and find that long-distance connections in particular cause a ``speed-up'' in temporal processing on a narrative alignment task.\\  

Within the Structured RC umbrella, numerous other \textit{brain-inspired} approaches have been proposed. In \cite{Damicelli2021-dm}, for example, the authors construct their reservoirs by ``scaling up'' small (approximately 30-neuron) cross-sections of macaque, marmoset, and human brain tissue, respectively; with their RC models they are able to achieve similar performance on memory capacity and sequence memory tasks compared to a vanilla RC. In another recent publication, \cite{Suarez2021-sm} uses patterns of human brain activity from diffusion-weighted imaging to create a functional 1000-node reservoir; among other findings, the authors report that their brain-inspired model has a lower ``wiring cost'' compared to the random RC.\\


In line with these works, herein we adopt a two-part strategy for imposing \textbf{explicit} brain structure onto the rate-based flavour of RCs, the \textit{Echo State Network} (see Fig. \ref{esn}). By ``explicit'', we refer to a one-to-one mapping of neurons onto an RC reservoir; this is distinct from the \textit{scaled up} and \textit{averaged} mappings in \cite{Damicelli2021-dm} and \cite{Suarez2021-sm}, respectively. In the first part of our approach, we draw from a completely-mapped brain region of interest (ROI) to create our derived reservoir. In the second part, we extract topological features from the reservoir in order to determine their -- and the original reservoir's -- impact on (predictive) performance. We use an early principles, mesoscale brain structure: from the \textit{smallest} animal which is behaviourally most-capable (i.e. in learning from and responding to external stimuli) and which also has a near-fully-mapped \textit{connectome} -- where ``neurons'' are simple point neurons, and ``weights'' are synapse affinities (i.e. the number of synaptic connections); here we refer to the \textit{olfactory region} of the \textit{fruit fly connectome}.
\\

The fruit fly is an \textit{excellent olfactory learner}: it (males specifically) can deduce the suitability of other flies for mating (i.e. their fertility) based on odor cues alone \cite{Hu2015-pc}; the fly can also be trained to increase or decrease its neurally-traceable output response to an odor gradient when the odor is followed with pleasure (i.e. food) or pain (i.e. electric shock), respectively \cite{Zhao2021-ng} -- impressively, it often learns to master these tasks in ``one-shot'' or ``few-shots'' \cite{Nowotny2014-bp, Dasgupta2017-pc, Zhao2021-ng}. Fruit fly olfactory-inspired algorithms, also, have successfully transferred many of these proficiencies to machine learning applications: \cite{Dasgupta2017-pc} for example, use the architecture of the ORN-to-MB (olfactory receptor neurons to mushroom body) portion of the olfactory system to create a locality-sensitive hashing algorithm -- here the authors report numerous improvements to various similarity search benchmarks. There is further motivation from the hypothesis that biological brains have been ``sharpened'' over time by natural selection to accurately and efficiently predict specific output responses from sensory inputs, which increase pleasurable outcomes and avoid painful ones -- in the fly brain, these have been shown to follow a gradient of prediction errors \cite{Bennett2021-km, Zhao2021-ng}. Whilst such ideas are beyond the scope of this work, they have served as a driving factor in our overall approach.
\\  
\subsection{Hypothesis}

We will investigate the contribution of explicit connectome-derived topology and also \textit{four connectome-derived structural features} -- each imposed onto a reservoir -- on the \textit{performance} and \textit{variance} of an Echo State Network (ESN) for size-variants of the Mackey-Glass, Lorenz, and Rossler time series. For structural features we will consider \textit{sparsity} (or density) of edges, the distribution of weights, weight positions (i.e. the set of all synaptic partners), and global clustering. We will measure \textit{performance} as the Mean-squared Error (MSE) -- see Eq. \ref{eqn: mse} -- between model predictions and output labels on a particular validation set; we will measure \textit{variance} by computing the squared standard deviation of MSE over all validation sets for a particular training input size. MSE is commonly used for ESN benchmarking -- i.e. in \cite{Racca2021-lz, Couillet2016-ie, Alao2021-hd}.

\begin{equation}\label{eqn: mse}
 E\bigg(\hat{\mathbf{y}}(n),\mathbf{y}(n)\bigg)=\sum_{n=1}^{T}\frac{1}{T}\bigg( \hat{\mathbf{y}}(n) - \mathbf{y}(n)\bigg)^2    
\end{equation}

Here $\hat{\mathbf{y}}(n)$ is an output prediction at discrete time step $n$, $\mathbf{y}(n)$ is a provided target label at step $n$, and $T$ is the total number of discrete time steps considered. 
\\

\subsection{Objectives}
From our hypothesis we derive \textit{five objectives}. Each objective is concerned with \textit{determining the impact of a particular connectome-derived feature} (or all features) on ESN \textit{performance and variance} -- by ESN, here we refer to a conventional baseline (see Methodology). For our first objective, we impose the full connectome ROI onto an ESN reservoir and measure the resulting time series prediction performance and variance. Our second objective explores the influence of connectome-derived \textit{sparsity} (or equivalently, \textit{density}). We define ``density'' as the proportion of all nonzero edges; for example, a 35\%-dense network of 10 neurons has 35 nonzero weights. Our third objective is to identify the effect of altering the \textit{parent distribution} of reservoir edge weights. For our fourth objective, we permute the set of all node pairs -- i.e. the ``position'' of edge weights. Finally, we measure the changes in model performance and variance as a result of adjusting the global \textit{clustering coefficient} (see Eq. \ref{eqn: C}) of the network, $C$; this is a descriptive term for small-world networks -- e.g. the fly brain \cite{Liao2017-ra, Porter2012-dr, Kaiser2015-rt}. 

\begin{equation}\label{eqn: C}
    C = 3 \times \frac{\text{\# of triangles}}{\text{\# of connected triples}}
\end{equation}

Here a ``connected triple'' is any three nodes $\{D,E,F\}$ where $\{D,E\}$ and $\{E,F\}$ are connected by two edges. For a triple to be \textit{closed} it requires the nodes to be connected by three edges. A \textit{triangle} graph holds three closed triples.
\\

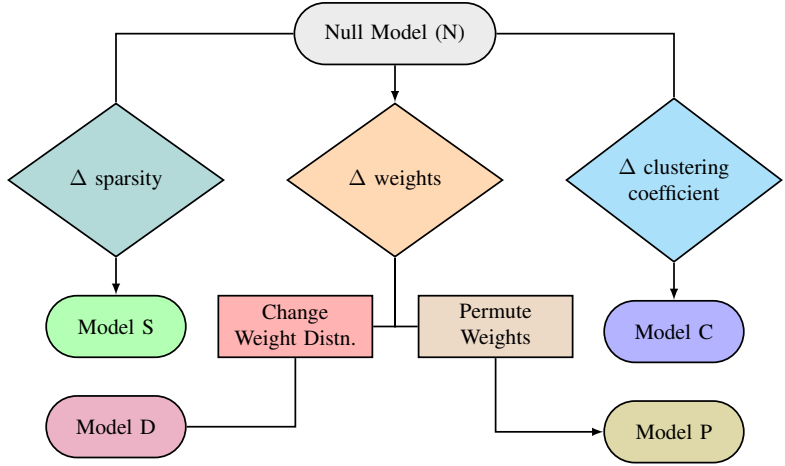
\begin{figure*}[h]
\centering
\resizebox{10.5cm}{!}{
\begin{tikzpicture}[font=\normalsize,thick]

\node[draw,
    align=center,
    rounded rectangle,
    minimum width=3.5cm,
    fill=lightgray!30,
    minimum height=1cm] (block1) {Null Model (N)};

\node[draw,
    diamond,
    below=0.6cm of block1,
    minimum width=3.5cm,
    fill=orange!30,
    minimum height=2.5cm,
    inner sep=0] (block2) { $\Delta$ weights};

\node[draw,
    diamond,
    left=of block2,
    minimum width=3.5cm,
    fill=teal!30,
    minimum height=2.5cm,
    inner sep=0] (block3) { $\Delta$ sparsity};

\node[draw,
    diamond,
    text width=2cm,
    align=center,
    right=of block2,
    minimum width=3.5cm,
    fill=cyan!30,
    minimum height=2cm,
    inner sep=0] (block4) { $\Delta$ clustering coefficient};

\node[draw,
    rounded rectangle,
    minimum width=2.5cm,
    below=0.6cm of block3,
    fill=green!30,
    minimum height=1cm] (block5) {Model S};

\node[draw,
    rectangle,
    right=0.5cm of block5,
    text width=2cm,
    align=center,
    fill=red!30,
    minimum width=2.5cm,
    minimum height=1cm,
    inner sep=0] (block6) {Change Weight Distn.};

\node[draw,
    rectangle,
    below=of block2,
    align=center,
    text width=2cm,
    fill=brown!30,
    right=3.75cm of block5,
    minimum width=2.5cm,
    minimum height=1cm,
    inner sep=0] (block7) {Permute Weights};

\node[draw,
    rounded rectangle,
    minimum width=2.5cm,
    below=0.6cm of block4,
    fill=blue!30,
    minimum height=1cm] (block8) {Model C};

\node[draw,
    rounded rectangle,
    minimum width=2.5cm,
    fill=purple!30,
    below=0.6cm of block5,
    minimum height=1cm] (block9) {Model D};

\node[draw,
    rounded rectangle,
    minimum width=2.5cm,
    fill=olive!30,
    below=0.6cm of block8,
    minimum height=1cm] (block10) {Model P};

\draw[-latex] (block1) edge (block2)
    (block1) -| (block3)
    (block1) -| (block4)
    (block4) edge (block8)
    (block3) edge (block5)
    (block2) |- (block6)
    (block2) |- (block7)
    (block6) |- (block9)
    (block7) |- (block10);
\end{tikzpicture}
}
\caption{Model selection based on modifications to a null model; ``Distn.'' is short for ``Distribution''.} \label{models}
\end{figure*}

To address our hypothesis and implement our objectives, we propose a \textit{subtractive} model-driven framework (Fig. \ref{models}). We start with a (biological) \textit{null model}: an ESN with its reservoir entirely replaced by a fruit fly connectome (i.e. its connectivity matrix). We then ``remove'' connectome topological features, replace these with conventional ESN analogues, and determine the resulting impact on performance and variance for time series prediction of subsets of the Mackey-Glass, Lorenz, and Rossler time series. We construct four models: For Model S (``Sparsity''), we increase the \textit{density} of the null model (from 1.3\%) to 20\% -- as in \cite{noauthor_undated-uq,Gallicchio2020-fl, Lukosevicius2009-hr}. To fill zero-weight positions (selected randomly) with reservoir weights, we sample from a bootstrapped population distribution of nonzero connectome weights. For Model D (``Distribution''), we switch each nonzero null model weight with one sampled from a uniform distribution on $[-1,1]$ -- as in \cite{noauthor_undated-uq, Wu2018-ey}. For Model P (``Position''), we permute all nonzero null model weights from the connectivity matrix row-wise and column-wise, whilst retaining zero-valued positions. For Model C (``Clustering''), we add bootstrapped connectome weights to the null model to increase $C$ from 0.27 (for our connectome) to 0.5 -- i.e. greater than that of the conventional (random) ESN reservoir.\\

\section{Related work}\vspace{.5cm}
\subsection{Structural features of Echo State Networks}

\textit{Sparsity}: In \cite{Lukosevicius2009-hr} and \cite{Gallicchio2020-fl}, a sparsely-connected reservoir -- of 20\% density -- is recommended for optimal performance; this enables sufficient short-term memory capacity of the network. \cite{noauthor_undated-uq} considers a 20\%-density ESN by default. \cite{Lukosevicius2012-rq} recommends 10 edges per node on average as a ``rule of thumb''. Conversely, \cite{Ozturk2007-lv} finds that varying density from 7\% to 20\% has a negligible effect on average state entropy -- i.e. the richness of the space of input representations.\\  

\noindent \textit{Weights}: ESN reservoir weights are fixed, random values from a uniform distribution on $[-1,1]$, by convention \cite{Lukosevicius2009-hr, Gallicchio2020-fl}. Some works have sought to incorporate dynamic weights: \cite{Tortorella2021-ls}, for example, found that a dynamic graph ESN could perform twelve classification tasks with similar accuracy to non-dynamic graph ESNs using less memory. The authors of \cite{Wu2018-ey} sample reservoir weights from uniform, arcsine, and gaussian distributions: they find that an arcsine distribution can improve time series prediction performance, but speculate that it is the sparsity resulting from the distribution (and not the sampled values themselves) that is responsible. \cite{Gallicchio2020-fl}, in particular, recommends an investigation into the behaviour of reservoir topologies with structured (i.e. non-random) weights.\\

\noindent \textit{Small-worldness}: A small-world network exhibits sufficiently slow increase in mean shortest path length (MSPL) with the addition of new nodes \cite{Porter2012-dr}. Alternatively, a network is \textit{small-world} in comparison to an Erdos-Renyi random graph if it has lower MSPL and higher clustering coefficient ($C$) \cite{Karyotis2016-ou}; $C$ and MSPL are both important features to determine small-worldness \cite{Liao2017-ra, Porter2012-dr}. \cite{Kawai2017-ez} observe similar time series prediction performance with fewer reservoir neurons when the reservoir -- after node removal -- is small-world (in addition to other constraints). \\

\subsection{Fly-inspired machine learning approaches}
\cite{Tschopp2018-iu} apply a connectome-derived approach to motion detection. In their work, the authors create a Recurrent Convolutional Neural Network (RCNN) from the early anatomical stages of fruit fly vision -- the Elementary Motion Detection (EMD) circuit, from retinal mapping of inputs to the T4 and T5 neurons at the lobula plate (ON/OFF pathway); \cite{Tschopp2018-iu} impart this topology explicitly and also include all known excitatory and inhibitory synapses. Their RCNN convolutional layers form a hexagonal lattice structure; cells in each lattice correspond to biological neurons, and each layer represents an appropriate anatomical layer (i.e. multiple parallel layers represent the retinal cells). \cite{Tschopp2018-iu} initially train their network on a subset of DAVIS 2016 -- on video clips where objects move across a fixed lens. They compare validation performance versus an equivalent network with randomly-generated weights on an object detection task; specifically they move light and dark bars in 180 directions across the ``visual field'' of each network. Surprisingly, their RCNN is able to capture the same directional and orientation selectivity which is observed in the fly brain, whereas the randomly-weighted network is not. This suggests that fly-like functionalities can be recovered from a connectome alone. In contrast to \cite{Tschopp2018-iu}'s explicit approach, \cite{Liang2021-ej} take ``loose inspiration'' from the principles of sparse coding observed in the mushroom body Kenyon cells and construct a corresponding network to learn semantic representations of words and to generate word embeddings from an unstructured text corpus; they report comparable performance to conventional techniques (BERT, GloVe) whilst using less memory and training time. In our own previous work, we imposed fruit fly connectome-derived weights onto an ESN reservoir for Mackey-Glass time series prediction on 300 and 900 training inputs \cite{Morra2022-ml}. We observed a significant reduction in model variance and an improvement in predictive performance; however, it was unclear from our results whether the \textit{entire} topology was responsible for our results, or if \textit{particular structural} features could be isolated to provide equivalent (or greater) performance or variance benefits. Furthermore, we were not aware of whether our results would generalize across multiple datasets. These aspects have motivated our current set of experiments.

\vspace{.2cm}
\section{Methodology}\vspace{.2cm}
\subsection{Building a connectivity matrix}

We derive all null model weights from the \textit{hemibrain}: a connectome of an adult (female) fruit fly, comprising 21,734 uncropped (i.e. containing most arbors) and 4,456 cropped neurons, and over 20 million traced synapses \cite{Scheffer2020-uc}. We query the hemibrain through its publicly-available API \cite{Scheffer2020-uc}; in particular, selecting all neurons in the olfactory system major ROIs (regions of interest) -- the antennal lobe, mushroom body, and lateral horn -- and all of their presynaptic and postsynaptic partners; we then store all neuron-neuron pairs by body ID in addition to their edge weights -- equivalent to the number of synaptic connections \cite{Scheffer2020-uc}; finally, we construct an adjacency (connectivity) matrix from these pairs using the NetworkX package, where row and column labels correspond to particular neuron body IDs, and where each cell value represents the ``weight'' \textit{from} a particular row body ID \textit{to} a particular column body ID \cite{SciPyProceedings_11}. From here we enact one assumption and a simplification: First, as in \cite{Proske2012-yt, Gao2020-xm} we assume that each neuron is self-connected by filling the diagonal of the connectivity matrix $M$ with ones. Second, we truncate our connectome selection to include only the \textit{right lateral horn}; this holds the reservoir to a size of 4,286 neurons (instead of 17,421) and tens (instead of hundreds) of millions of weights, and ultimately reduces training times and model complexities. 

\vspace{.5cm}
\subsection{The Echo State Network}
We describe the network dynamics of the ESN from \cite{Lukosevicius2012-rq}. For some discrete time series input $\mathbf{u}(n) \in \mathbb{R}^{N_u}$ and known output $\mathbf{y}(n) \in \mathbb{R}^{N_u}$, the ESN learns a prediction signal $\hat{\mathbf{y}}(n) \in \mathbb{R}^{N_y}$ which minimizes $E\bigg(\hat{\mathbf{y}}(n), \mathbf{y}(n)\bigg)$. The recurrent \textit{reservoir} layer of the ESN transforms the input time series $\mathbf{u}(n)$ as illustrated in Eq. \ref{eqn:2}. The update equation for the reservoir is provided in Eq. \ref{eqn:3}; and depends on the \textit{leaking rate} $\alpha$. $\alpha \in (0,1]$ controls the reservoir update speed.

\begin{equation}{\label{eqn:2}}
    \tilde{\mathbf{x}}(n) = \mathit{tanh}\bigg(\mathbf{W}^{in}[1;\mathbf{u}(n)]+\mathbf{Wx}(n-1)\bigg)
\end{equation}

\begin{equation}{\label{eqn:3}}
    \mathbf{x}(n) = (1-\alpha)\mathbf{x}(n-1)+\alpha \tilde{\mathbf{x}}(n)
\end{equation}

From Eq. \ref{eqn:2}, $\tilde{\mathbf{x}}(n)$ is the reservoir activation for time step $n$, $\mathit{tanh}$ is the network activation function, $\mathbf{W}^{in} \in \mathbb{R}^{N_x \times (1+N_u)}$ is the input-to-reservoir weight vector, $[\cdot;\cdot]$ represents a column vector, $\mathbf{W} \in \mathbb{R}^{N_x \times N_x}$ is the reservoir-to-reservoir weight vector, and $\mathbf{x}(n-1)$ is the previous reservoir activation.\\

The computation of the ESN's predicted output time series $\hat{\mathbf{y}}(n)$ is described in Eq. \ref{eqn:4}; and equivalently in Eq. \ref{eqn:5} (matrix notation). In Eq. \ref{eqn:4}, $\mathbf{W}^{out} \in \mathbb{R}^{N_y \times (1+N_u+N_x)}$ is the reservoir-to-output weight vector. In Eq. \ref{eqn:5}, $\hat{\mathbf{Y}} \in \mathbb{R}^{N_y \times T}$ -- where $\mathit{T}$ is the length of the input time series (not to be confused with the transpose in Eq. \ref{eqn:6}) -- includes all predictions $\hat{\mathbf{y}}(n)$, and $\mathbf{X} \in \mathbb{R}^{(1+N_u+N_x) \times T}$ is the design matrix which includes all column vectors $[\mathbf{1};\mathbf{U};\mathbf{X}]$ from $[1;\mathbf{u}(n);\mathbf{x}(n)]$.

\begin{equation}{\label{eqn:4}}
    \hat{\mathbf{y}}(n)=\mathbf{W}^{out}[1;\mathbf{u}(n);\mathbf{x}(n)]
\end{equation}
\begin{equation}{\label{eqn:5}}
    \hat{\mathbf{Y}} = \mathbf{W}^{out}\mathbf{X}
\end{equation}

Consider Eq. \ref{eqn:5.5} for finding optimal weights $\mathbf{W}^{out}$. To solve this system we use Ridge Regression (regression with L2 regularization). Here $\mathbf{Y} \in \mathbb{R}^{N_y \times T}$ is the known time series output in matrix form, $\lambda$ is the \textit{regularization coefficient}, and $\mathbf{I}$ is the identity matrix \cite{Lukosevicius2012-rq}.

\begin{equation}{\label{eqn:5.5}}
    \mathbf{Y} = \mathbf{W}^{out}\mathbf{X}
\end{equation}
\begin{equation}{\label{eqn:6}} \mathbf{W}^{out}=\mathbf{Y}\mathbf{X}^{T}\bigg(\mathbf{X}\mathbf{X}^{T}+\lambda\mathbf{I}\bigg)^{-1}
\end{equation}

\subsection{Time series prediction}
We consider the Mackey-Glass 17 (MG-17), Lorenz, and Rossler time series, which are used conventionally for benchmarking ESNs. We observe that MG-17 is the \textit{most} commonly-used \cite{Lopez-Caraballo2016-kz, Gallicchio2020-fl, Wu2018-ey, Aceituno2020-lf, Hart2021-ep, Arroyo2020-op}; it is described by the following differential equation.

\begin{equation}\label{eqn:7}
\frac{dx}{dt}=\beta x(t)+\frac{\gamma x(t-\tau)}{1+x(t-\tau)^{10}}
\end{equation}

Here $[\beta, \gamma, \tau]$ are some fixed, real-valued parameters. $x(t)$ is the value of the time series at time $t$, delayed by time $\tau$ -- $\tau=17$ produces chaotic behaviour \cite{Lopez-Caraballo2016-kz}. The differential equations for the Lorenz and Rossler time series, respectively, are described below \cite{Weeks_undated-ug}:

\begin{equation}
\frac{dx}{dt}=\sigma (y-x),
\frac{dy}{dt}= rx-y-xz,
\frac{dx}{dt}= xy - bz,
\end{equation}

Here $\sigma = 10$, $r=28$, and $b=8/3$ are used.

\begin{equation}
\frac{dx}{dt}= -z-y,
\frac{dy}{dt}= x+ay,
\frac{dx}{dt}= b+z(x-c),
\end{equation}

Here $a=0.15$, $b=0.2$, and $c=10$ are used.
\\

We task all models to perform multi-step prediction on each time series. Specifically, for a time series segment of length $M \in \{2m: m \in \mathbb{Z}^{+}\}$ and given an input time series $u(n)$ for $n \in [0, ... ,\frac{M}{2}]$, each model must predict an output time series $\hat{y}(n)$ for $n \in [\frac{M}{2}+1, ... ,{M}]$. We consider subsets of the time series of length 600, 1200, 1800, and 2400 for MG-17 -- starting at time $n=0$ from \cite{noauthor_2016-jq} -- and 600 and 1800 for Lorenz and Rossler. These sequence lengths are comparable with other benchmarks \cite{noauthor_undated-uq, Lopez-Caraballo2016-kz, Ozturk2007-lv, Maat2019-vv}. We use an 80:20 train-validate split -- as in \cite{Gallicchio2020-fl} -- and denote the training input size (TR$\_$IN) for all experiments. We use EasyESN \cite{noauthor_undated-uq} to train and validate our models. Data for all analysis are found at \cite{noauthor_2016-jq} and \cite{Weeks_undated-ug}.\\

\subsection{Model preparation and comparison}
\noindent{\textit{Model creation}}: The models -- Null (N), S (``Sparsity''), D (``Distribution''), P (``Position''), and C (``Clustering'') -- have been constructed as previously specified (see Introduction). The conventional ESN model -- which, from here on we refer to as ``E'' -- has a reservoir comprised of (\textit{seeded}) random values from a uniform distribution on $[-1,1]$ \cite{Lukosevicius2009-hr, Gallicchio2020-fl}; this is in addition to all standard ESN trappings: for example, we use a spectral radius ($\rho$) of 1, which is advised to retain the Echo State Property (ESP) \cite{Venayagamoorthy2009-wu, Jaeger2007-bd, noauthor_undated-uq}. $\rho$ is conventionally divided by the maximum eigenvalue of the reservoir weight vector, and acts to alter the distribution of weights. We provide additional experiments with $\rho=0.3$ and $\rho=0.7$ -- these models are called ``E3'' and ``E7'', respectively. A final parameter of consideration is the transient time $\tau$. As a result of setting $x(0) = 0$ arbitrarily, the reservoir activation is in an ``unnatural starting state'', and so the first $\tau$ time steps need to be discarded \cite{Lukosevicius2012-rq}; we use $\tau$=100. In addition to the conventional and varying spectral radius models, we consider an additional conventional ESN with a new random seed -- we call this model ``ES'' (ESN seeded). Finally, we consider a ``distribution-matched'' ESN, where we map weights from $[-1,1]$ to $[0, max(W)]$, where $W$ is the set of all connectome reservoir weights; namely, we apply a scaling factor after taking the magnitude of all weights -- we call this model ``ED''.
\\

\noindent{\textit{Model selection}}: We select optimal hyperparameters (${\alpha, \lambda}$) for each model class using a grid search: we train and validate models on size-variants of the Mackey-Glass, Lorenz, and Rossler time series and report those hyperparameters which yield the best performance (MSE) on the validation set. For the grid search we consider a small subset of possible values within the specified parameter ranges (35 total). For $\alpha$ we consider a range of $[0,1]$ by convention \cite{Lukosevicius2012-rq}. For $\lambda$ we select in $[0,1E+03]$. 
\\

\noindent{\textit{Model evaluation}}: We create the best instances from each model class with hyperparameters as discovered in the model selection step. We then train and validate each instance 30 times on the Mackey-Glass, Lorenz, and Rossler time series for training input (TR$\_$IN) sizes of $\{250, 500, 750, 1000\}$ and $\{250, 750\}$, respectively. We also conduct 30 trials of training and validation following hyperparameter optimization for experiments with additional models (ED, ES, E3 and E7). Where applicable, significant differences are computed using the Wilcoxon signed ranked test ($p < 0.05$). All model selection and evaluation is conducted on an ``e2-highcpu-8'' Google Cloud instance with 8 vCPUs (1 GB memory per CPU) and 8 GB system RAM on an x86 platform -- Debian 11 (Bullseye).
\\

\section{Experiments}\vspace{.5cm}

\begin{figure*}[!h]
    \centering
    \includegraphics[width=0.65\textwidth]{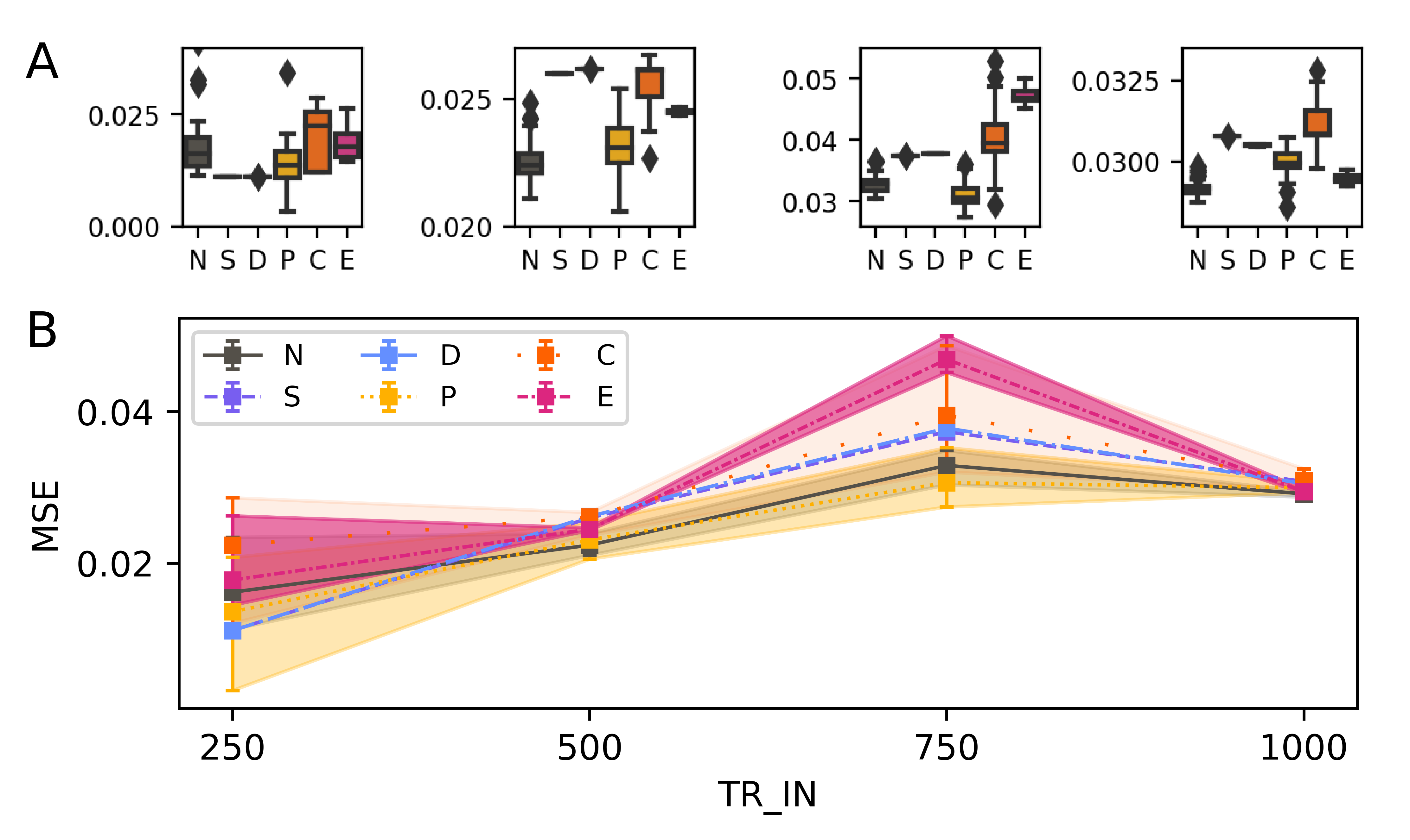}
    \caption{Validation scores across 30 trials on MG-17. A) From left to right, box plots are for MG-17 on TR\_IN = 250, 500, 750, and 1000 time steps. B) A line plot with error margins summarizing model results across TR\_IN sizes.}
    \label{fig: mgall}
\end{figure*}

\begin{figure}[!h]
    \centering
    \begin{center}
    \hspace{-1cm}
    \includegraphics[width=0.4\textwidth]{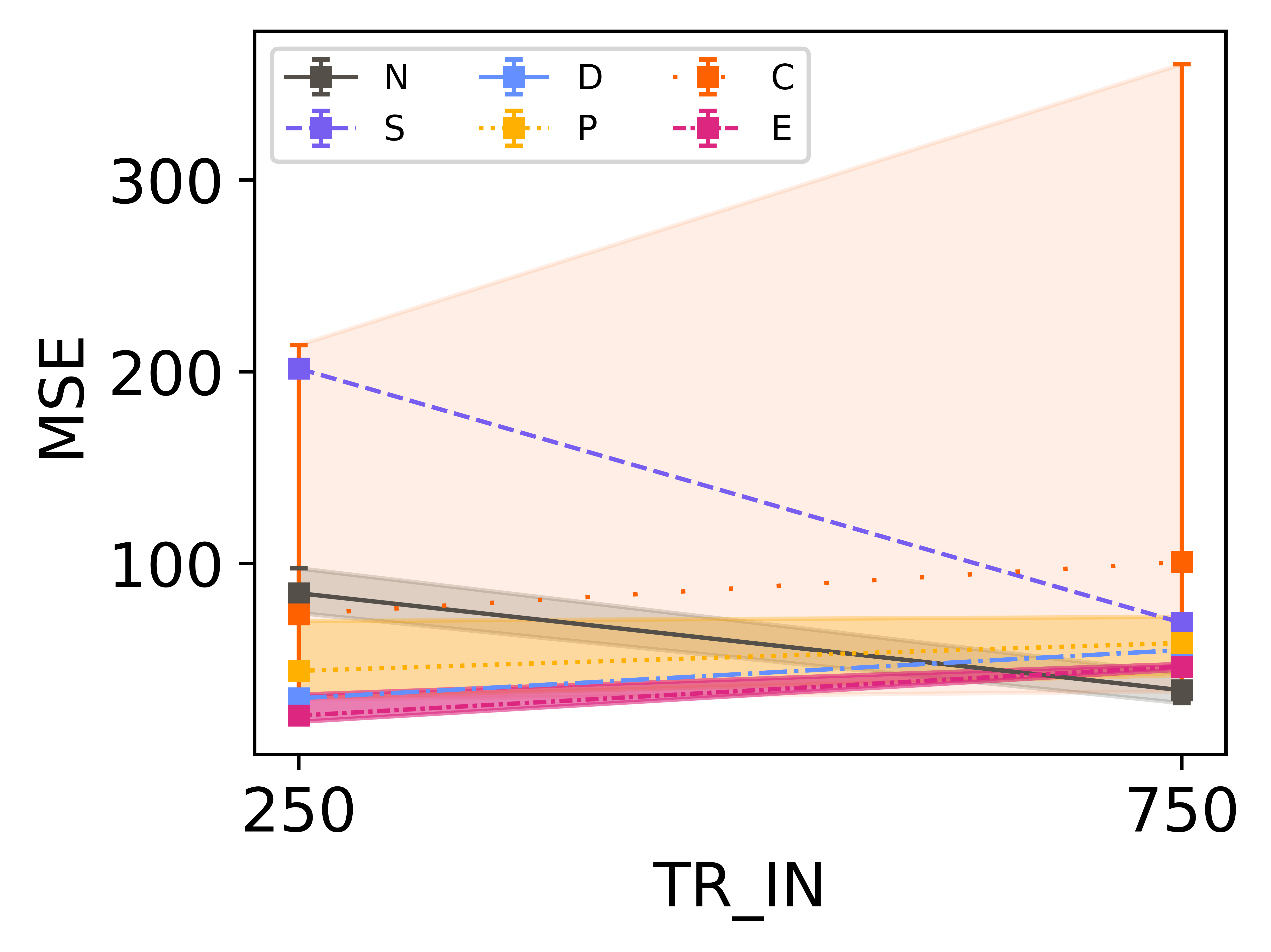}
    \end{center}
    \caption{Lorenz validation scores for TR\_IN = 250 and 750 discrete time steps for models N, S, D, P, C, and E.}
    \label{fig: lorenz250750}
\end{figure}

\begin{figure}[!h]
    \centering
    \hspace{-1cm}
    \includegraphics[width=0.4\textwidth]{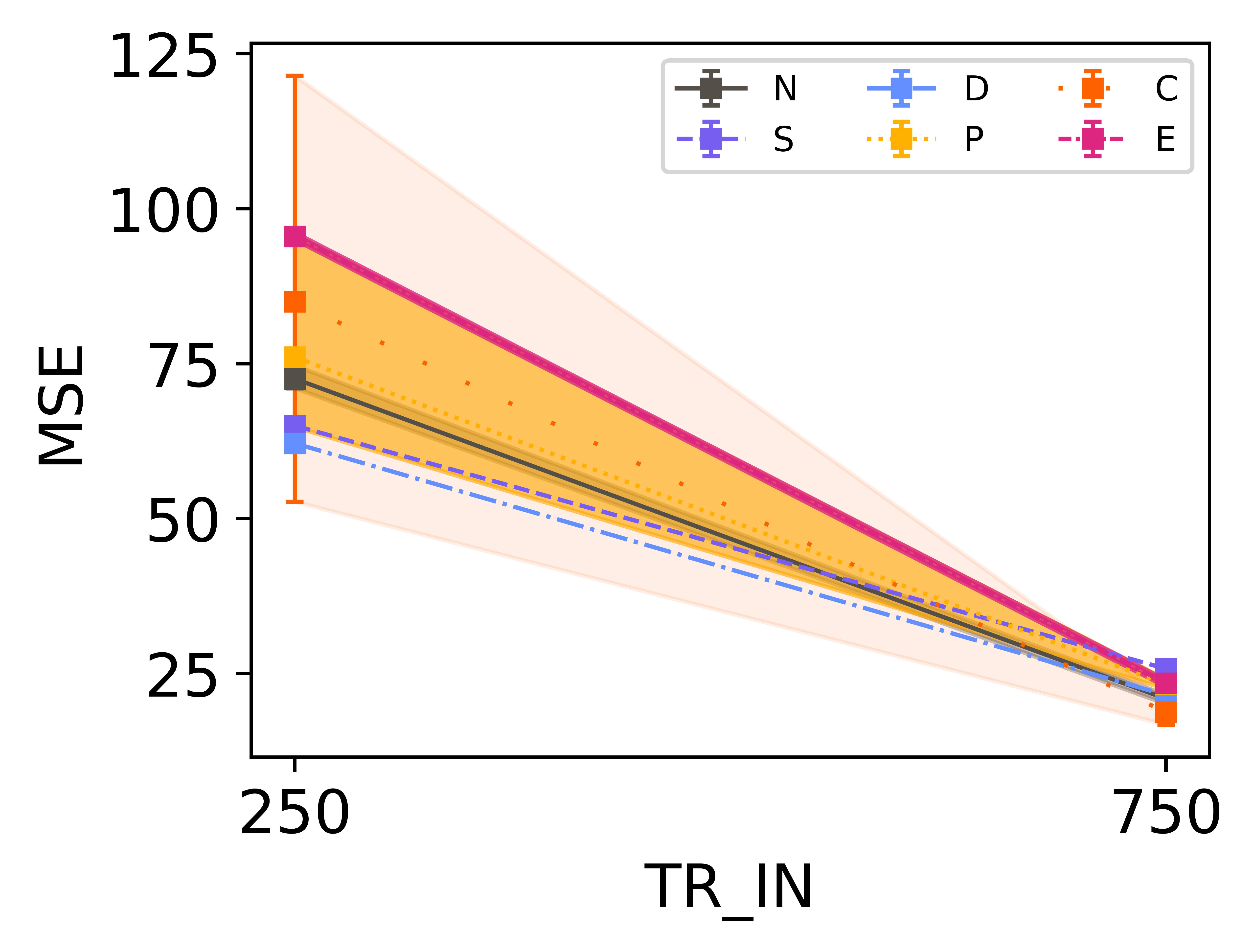}
    \caption{Validation scores for models N, S, D, P, C, and E on the Rossler chaotic time series for TR\_IN = 250 and 750.}
    \label{fig: rossler250750}
\end{figure}

\begin{table*}[!h]
\caption{Performance and Variance Comparisons between Models N, E, S, D, P, and C on MG-17}
\begin{center}
\begin{tabular}{@{}lllllllllll@{}}
\toprule
     & \multicolumn{2}{l}{MG${}_{\text{(TR}\text{\_}\text{IN}=250\text{)}}$}     & \multicolumn{2}{l}{MG${}_{\text{(TR}\text{\_}\text{IN}=500\text{)}}$}     & \multicolumn{2}{l}{MG${}_{\text{(TR}\text{\_}\text{IN}=750\text{)}}$}     & \multicolumn{2}{l}{MG${}_{\text{(TR}\text{\_}\text{IN}=1000\text{)}}$}   \\ 
     & MSE              & ${\sigma}^2$     & MSE              & ${\sigma}^2$     & MSE              & ${\sigma}^2$     & MSE              & ${\sigma}^2$  \\ \midrule
N & \textbf{1.6E-02} ${}_{\pm  4.3E-03}$ & 1.3E-04 & \textbf{2.2E-02} ${}_{\pm 3.2E-04}$  & 7.2E-07 & \textbf{3.3E-02} ${}_{\pm 5.7E-04}$  & 2.3E-06 & \textbf{2.91E-02} ${}_{\pm 9.3E-05}$  & 6.1E-08\\
E    & 1.8E-02 ${}_{\pm 1.2E-03}$ & 9.9E-06 & 2.4E-02 ${}_{\pm 2.8E-05}$  & 5.4E-09 & 4.7E-02 ${}_{\pm 4.4E-04}$  & 1.4E-06 & 2.95E-02 ${}_{\pm 4.9E-05}$  & 1.7E-08\\ \midrule
S    & 1.1E-02 ${}_{\pm 1.7E-18}$ & 2.7E-35 & 2.6E-02 ${}_{\pm 5.8E-11}$  & 2.4E-20 & 3.7E-02 ${}_{\pm 3.8E-15}$  & 1.0E-28 & 3.1E-02 ${}_{\pm 5.2E-13}$  & 1.9E-24 \\
D    & 1.1E-02 ${}_{\pm 8.8E-07}$ & 5.3E-12 & 2.6E-02 ${}_{\pm 1.7E-06}$  & 2.0E-11 & 3.8E-02 ${}_{\pm 3.2E-06}$  & 7.0E-11 & 3.0E-02 ${}_{\pm 5.9E-06}$  & 2.4E-10 \\
P   & 1.4E-02 ${}_{\pm 2.0E-03}$ & 2.9E-05 & 2.3E-02 ${}_{\pm 6.7E-04}$  & 3.1E-06 & 3.1E-02 ${}_{\pm 7.1E-04}$  & 3.5E-06 & 3.0E-02 ${}_{\pm 1.7E-04}$  & 2.0E-07  \\
C    & 2.2E-02 ${}_{\pm 2.5E-03}$ & 4.2E-05 & 2.6E-02 ${}_{\pm 3.6E-04}$  & 8.8E-07 & 4.0E-02 ${}_{\pm 1.9E-03}$  & 2.4E-05 & 3.1E-02 ${}_{\pm 2.5E-04}$  & 4.5E-07\\ \bottomrule
\end{tabular}
\end{center}
\label{Tab:tab1}
\end{table*}


\begin{table*}[]
\caption{Performance and Variance Comparisons between Models N, E, S, D, P, and C on Lorenz and Rossler}
\begin{center}
\begin{tabular}{@{}lllllllll@{}}
\toprule
     & \multicolumn{2}{l}{L${}_{\text{(TR}\text{\_}\text{IN}=250\text{)}}$}      & \multicolumn{2}{l}{L${}_{\text{(TR}\text{\_}\text{IN}=750\text{)}}$}      & \multicolumn{2}{l}{R${}_{\text{(TR}\text{\_}\text{IN}=250\text{)}}$}      & \multicolumn{2}{l}{R${}_{\text{(TR}\text{\_}\text{IN}=750\text{)}}$}      \\
     & MSE              & ${\sigma}^2$     & MSE              & ${\sigma}^2$     & MSE              & ${\sigma}^2$     & MSE              & ${\sigma}^2$     \\ \midrule
N & 8.5E+01 ${}_{\pm 2.3E+00}$ & 3.7E+01 & \textbf{3.4E+01} ${}_{\pm 2.1E+00}$ & 3.0E+01 & \textbf{7.3E+01} ${}_{\pm 3.6E-01}$ & 9.0E-01 & \textbf{2.1E+01} ${}_{\pm 1.7E-01}$ & 2.0E-01 \\
E    & \textbf{2.1E+01} ${}_{\pm 2.0E+00}$ & 2.8E+01 & 4.6E+01 ${}_{\pm 2.9E-01}$ & 5.9E-01 & 9.6E+01 ${}_{\pm 1.1E-01}$ & 9.1E-02 & 2.3E+01 ${}_{\pm 1.0E-01}$ & 7.1E-02 \\ \midrule
S    & 2.0E+02 ${}_{\pm 6.8E-11}$ & 3.2E-20 & 6.9E+01 ${}_{\pm 0.0E+00}$     & 0.0E+00 & 6.5E+01 ${}_{\pm 4.5E-07}$ & 1.4E-12 & 2.6E+01 ${}_{\pm 5.1E-07}$ & 1.8E-12 \\
D    & 3.0E+01 ${}_{\pm 1.4E-01}$ & 1.3E-01 & 5.5E+01 ${}_{\pm 1.1E-01}$ & 8.4E-02 & 6.2E+01 ${}_{\pm 5.6E-03}$ & 2.2E-04 & 2.2E+01 ${}_{\pm 8.3E-03}$ & 4.8E-04 \\
P   & 4.4E+01 ${}_{\pm 4.7E+00}$ & 1.5E+02 & 5.9E+01 ${}_{\pm 1.3E+01}$ & 1.1E+03 & 7.6E+01 ${}_{\pm 4.9E+00}$ & 1.6E+02 & 2.3E+01 ${}_{\pm 1.4E-01}$ & 1.5E-01 \\
C    & 7.4E+01 ${}_{\pm 1.3E+02}$ & 1.1E+05 & 1.0E+02 ${}_{\pm 6.2E+01}$ & 2.7E+04 & 8.5E+01 ${}_{\pm 5.9E+00}$ & 2.4E+02 & 1.9E+01 ${}_{\pm 4.0E-01}$ & 1.1E+00 \\ \bottomrule
\end{tabular}
\end{center}
\label{Tab:tab2}
\end{table*}


\begin{table*}[!h]
\caption{Performance and Variance Comparisons between N and Random Alternatives (ED, ES, E3, and E7)}
\begin{center}
\begin{tabular}{@{}lllllll@{}}
\toprule
      &     & N                   & ED                 & ES            & E3                & E7                \\ \midrule
MG${}_{\text{(TR}\text{\_}\text{IN}=250\text{)}}$ & MSE & 1.6E-02 ${}_{\pm  4.3E-03}$  & \textbf{1.1E-02} ${}_{\pm  1.7E-18}$  & 2.1E-02 ${}_{\pm 1.6E-03}$  & 1.2E-02 ${}_{\pm  3.5E-06}$  & 1.6E-02 ${}_{\pm  9.9E-05}$  \\
      & ${\sigma}^2$ & 1.3E-04             & 1.2E-35             & 1.7E-05             & 8.7E-11             & 6.8E-08             \\ \midrule
MG${}_{\text{(TR}\text{\_}\text{IN}=750\text{)}}$ & MSE & \textbf{3.3E-02} ${}_{\pm  5.7E-04}$  & 3.7E-02 ${}_{\pm  0.0E+00}$  & 4.7E-02 ${}_{\pm  2.0E-04}$  & 3.8E-02 ${}_{\pm  1.5E-05}$  & 4.3E-02 ${}_{\pm  8.8E-05}$  \\
      & ${\sigma}^2$ & 2.3E-06             & 4.8E-35             & 2.8E-07             & 1.6E-09             & 5.4E-08             \\ \bottomrule
\end{tabular}
\end{center}
\label{Tab:tab3}
\end{table*}


\begin{table*}[!h]
\caption{Additional Performance and Variance Comparisons between N and ED}
\begin{center}
\begin{tabular}{@{}lllllll@{}}
\toprule
      &     & N                   & ED               \\ \midrule
MG${}_{\text{(TR}\text{\_}\text{IN}=500\text{)}}$ & MSE & \textbf{2.2E-02} ${}_{\pm  3.2E-04}$  & 2.6E-02 ${}_{\pm  5.5E-11}$  \\
      & ${\sigma}^2$ & 7.2E-07             & 2.1E-20  \\ \midrule
MG${}_{\text{(TR}\text{\_}\text{IN}=1000\text{)}}$ & MSE & \textbf{2.9E-02} ${}_{\pm  9.3E-05}$  & 3.1E-02 ${}_{\pm  5.5E-13}$    \\
      & ${\sigma}^2$ & 6.1E-08             & 2.1E-24 \\ \bottomrule
\end{tabular}
\label{Tab:tab4}
\end{center}
\end{table*}

\subsection{Mackey-Glass 17 chaotic time series}
Fig. \ref{fig: mgall}B provides a comprehensive illustration of all MG-17 prediction results for TR\_IN values of 250, 500, 750, and 1000; Fig. \ref{fig: mgall}A provides a particular ``snapshot'' at each TR\_IN value in ascending order. From these plots we observe a consistent improvement in predictive performance from the conventional model (E) to the null model (N). For TR\_IN = 750 the difference is prominent: here we observe a $\approx$ 30\% reduction in error. Regarding models S, D, P and C, we observe high variance for models P and C through all training input sizes; at TR\_IN = 750, for example, Model P yields an approximately 2-fold increase in variance over the null model, whereas Model C exhibits a 10-fold increase. Model S and Model D show a dramatic reduction in variance compared to the null model for all training input sizes considered -- however, we note that their overall performance in relation to other models across training input sizes varies; they both, for example, yield the highest MSE values at TR\_IN = 500 and the lowest for TR\_IN = 250. Model results are summarized in Table \ref{Tab:tab1}; which also highlights differences between the null and conventional models -- bold values indicate the (significant) winner for a particular TR\_IN value.\\

\subsection{Lorenz chaotic time series}
In Fig. \ref{fig: lorenz250750} we observe that the conventional ESN outperforms the null model -- and all other models -- for 250 training input \\

\noindent steps. We also observe that the null model outperforms the conventional ESN and all other models as the dataset size increases to TR\_IN = 750. For Model C, in particular, we observe an error margin which extends beyond the range of reported MSE values for all other models; and looking at Table \ref{Tab:tab2}, we see for TR\_IN = 250 that the variance for Model C is an approximately 1000-fold increase over the model which yields the next-largest variance (P). Model P, in turn, has a 4 times variance increase compared to the null model. \\

\subsection{Rossler chaotic time series}
In Fig. \ref{fig: rossler250750} we observe that the null model outperforms the conventional ESN in both trials considered (TR\_IN = 250 and TR\_IN = 750). We also note that Models S and D outperform all others for TR\_IN = 250 while concurrently yielding the lowest variances by multiplicative factors of $6E+12$ and $4E+03$ over the null model, respectively. Conversely, Models P and C exhibit the highest variance of all models; C, in particular, yields an error margin which extends beyond the range of errors across all other models. Interestingly, the Rossler time series task is the only one on which \textit{all models} achieve performance improvements with increasing TR\_IN.\\

\subsection{Alternative Random Models}
Comparing the null model (N) to a newly-seeded conventional ESN (ES), we observe in Table \ref{Tab:tab3} that performance on MG-17 improves from ES to N with increasing TR\_IN. Looking at all other null and alternative model results, we see for TR\_IN = 750 that the null model achieves the best performance; however, at TR\_IN = 250, Model ED outperforms the null model. Additional experiments for models ED and N (Table \ref{Tab:tab4}) show that N maintains a performance lead over ED for TR\_IN values of 500 and 1000.\\

\noindent

\section{Conclusion}\vspace{.5cm}
We present \textbf{two major findings}: First, that \textit{explicit connectome topology} \textit{informs significant} \textit{improvements} in chaotic time series prediction performance across multiple benchmark datasets when compared to a conventional Echo State Network model of the same reservoir size; Second, that imposing small-worldness onto an ESN reservoir can dramatically increase model variance -- and that this holds true, albeit to a lesser extent, with non-connectome-derived weight positioning. \vspace{.5cm}



\section{Broader impact}\vspace{.5cm}
\subsection{Contribution}
This work suggests that biologically-imposed structural connectivity is well-suited for learning input-output representations; in particular, with respect to time-varying, chaotic signals, and through the lens of a reservoir computer. Most surprising is that this suitability is captured with \textit{structure alone}, independent from any application of biologically-motivated function; we, for example, did not consider excitatory or inhibitory synapses.
\\
 
\subsection{Limitations}
Although we used a near-complete connectome ROI, here we opted out of using a complete mapping from olfactory input to output in the fly. While it is suggested from our model comparisons that brain connectivity, broadly, is preferable to a random graph structure for chaotic time series prediction, it is still unclear whether some aspect of the fly is actually being recreated ``in silica''; to develop a better understanding of our models context within this larger question, next steps will involve implementing an end-to-end olfactory connectome via a Reservoir Computer.\vspace{1cm}

\bibliography{bib}
\end{document}